\documentclass{article}

\usepackage[preprint]{neurips_2023}

\usepackage[utf8]{inputenc} 
\usepackage[T1]{fontenc}    
\usepackage{hyperref}       
\usepackage{url}            
\usepackage{booktabs}       
\usepackage{amsfonts}       
\usepackage{amsmath}
\usepackage{nicefrac}       
\usepackage{microtype}      
\usepackage{makecell}       
\usepackage{xcolor}         
\usepackage{tikz}           
\usepackage{tikzscale}
\usetikzlibrary{positioning, fit}
\usepackage{subfig}
\usepackage{algorithm,algpseudocode}
\usepackage[normalem]{ulem} 
\usepackage{enumitem}  
\usepackage{wrapfig}
\usepackage{selectp} 

\usepackage{amsthm,amssymb}
\usepackage{cleveref}
\crefname{ineq}{inequality}{inequalities}
\crefname{equation}{Eq.}{Eq.}
\creflabelformat{ineq}{#2{\upshape(#1)}#3} 
\crefname{theorem}{Theorem}{Theorem}
\crefname{claim}{Claim}{Claim}
\crefname{lemma}{Lemma}{Lemma}
\crefname{appendix}{Appendix}{Appendices}
\crefname{figure}{Fig.}{Fig.}
\crefname{table}{Table}{Tables}
\crefname{section}{Sec.}{Sec.}
\crefname{algorithm}{Algorithm}{Algorithms}

\newtheorem{definition}{Definition}

\definecolor{baseline_color}{gray}{0.5}

\title{Optimizing Job Allocation using Reinforcement Learning with Graph Neural Networks}

\newcommand{\commentOut}[1]{\ignorespaces}

\author{%
  Lars C.P.M. Quaedvlieg\\
  EPFL, Switzerland \\
  \texttt{\{[first name].[last name]\}@epfl.ch} \\
}

\begin{document}

\maketitle

\begin{abstract}
    Efficient job allocation in complex scheduling problems poses significant challenges in real-world applications. In this report, we propose a novel approach that leverages the power of Reinforcement Learning (RL) and Graph Neural Networks (GNNs) to tackle the Job Allocation Problem (JAP). The JAP involves allocating a maximum set of jobs to available resources while considering several constraints. Our approach enables learning of adaptive policies through trial-and-error interactions with the environment while exploiting the graph-structured data of the problem. By leveraging RL, we eliminate the need for manual annotation, a major bottleneck in supervised learning approaches. Experimental evaluations on synthetic and real-world data demonstrate the effectiveness and generalizability of our proposed approach, outperforming baseline algorithms and showcasing its potential for optimizing job allocation in complex scheduling problems.
\end{abstract}

\section{Introduction}

Efficient job allocation plays a vital role in diverse fields, including healthcare, logistics, and manufacturing, enabling optimal utilization of resources and ensuring timely completion of tasks~\citep{owliya2012new, bash2007cool, kumar2018job, vijayakumar2022framework}. The Job Allocation Problem (JAP), which is introduced and investigated in this report, involves assigning a maximum number of jobs to available resources, taking into account various constraints. Traditional approaches to general scheduling~\citep{crama1997combinatorial, pinedo2012scheduling}, rooted in Combinatorial Optimization (CO), have provided valuable insights into the inherent challenges and trade-offs in solving allocation problems, but they often struggle to handle the increasing complexity and real-time dynamics of modern applications~\citep{wolsey1999integer, chaudhry2016research}.

To address these limitations, recent research has turned to machine learning techniques to learn effective schedules~\citep{zhang2020learning, zhang2022learning, park2021schedulenet, chen2021gnn}. However, the requirement of annotated data in supervised learning poses a problem for scheduling problems. Hence, it is currently infeasible to do supervised learning on NP-Hard problems~\citep{yehuda2020s}. 

Our approach showcases the use of reinforcement learning (RL) as a powerful technique to overcome the challenges posed by the requirement of annotated data. By learning directly from simulations and observing performance signals~\cite{kaelbling1996reinforcement}, our framework mitigates the need for extensive manual labeling. This enables us to leverage the benefits of RL for job allocation in a more practical and scalable manner.

Furthermore, we use graph neural networks to exploit the graph structure of the problem, as is commonly done in machine learning for CO problems~\citep{wang2020learning, gasse2019exact, schuetz2022combinatorial}.






To evaluate the performance of our proposed approach, we conducted experiments on both real-world and synthetic datasets. We compared our model with baseline algorithms. We demonstrate that the GNN consistently outperformed the baseline algorithms. Furthermore, out-of-distribution testing revealed the generalizability of the GNN. 

The remaining structure of the report is as follows: in~\cref{sec:related}, we give a background and discuss the most promising research related to our work. Then, in~\cref{sec:jsap}, we formalize the JAP. In \cref{sec:rl-formalism}, we discuss the RL framework used to approach the problem. Then, we describe the model architecture and the training process in~\cref{sec:methods}. Finally, we discuss experimental results in~\cref{sec:exp} and make conclusions in~\cref{sec:conc}.
\section{Related Works}
\label{sec:related}

The approach introduced in this report lies in the fields of combinatorial optimization, graph neural networks, and reinforcement learning. In this section, we discuss the most prominent research and explain its relation to our work.

\textbf{Combinatorial Optimization (CO)} has been a cornerstone in tackling complex allocation problems. Previous research has delved into various formulations and algorithms to optimize the allocation of limited resources efficiently. The exploration of heuristics~\citep{colorni1996heuristics, lorena1996relaxation}, approximation algorithms~\citep{vazirani2001approximation, turek1992approximate}, and exact methods~\citep{wolsey1999integer} has paved the way for understanding the inherent challenges and trade-offs in solving CO problems. In particular, studies addressing job allocation~\citep{penmatsa2006price}, nurse rostering~\citep{burke2004state}, and other scheduling problems~\citep{chaudhry2016research} have contributed valuable insights into the allocation domain.

In recent years, \textbf{Graph Neural Networks (GNNs)} have proven to be powerful tools for analyzing and learning on graph-structured data~\citep{zhou2020graph, wu2020comprehensive}. Advanced architectures such as the Graph Attention Network (GAT)~\citep{velickovic2018graph} and Graph Isomorphism Network (GIN)~\citep{xu2019powerful} have demonstrated remarkable capabilities in tackling node/edge-level and graph-level machine learning tasks, respectively. Importantly, the application of GNNs to combinatorial optimization has shown promise in effectively leveraging the inherent graph structure found in various problem domains~\citep{gasse2019exact, cappart2021combinatorial, schuetz2022combinatorial}, while also providing the benefit of invariance to the size of the graph. In this work, we utilize graph attention networks in this work since we are learning representations of edges.


\textbf{Reinforcement Learning (RL)} provides a principled framework for learning optimal policies through trial-and-error interactions with an environment. RL has demonstrated remarkable successes in various domains, from game-playing agents to robotic control~\citep{mahmud2018applications, polydoros2017survey}. When applied to scheduling problems, RL offers the potential to learn adaptive policies that optimize assignments and resource utilization~\citep{zhang2020learning, park2021schedulenet}.

As the JAP is a novel problem introduced in this report, there are no existing works directly addressing this problem. Nevertheless, previous studies have examined supervised learning approaches in other similar scheduling problems, by for example learning priority dispatch rules~\cite{chen2021gnn, ingimundardottir2011supervised} to compute allocation policies, but the requirement of manual labels presents a significant challenge. To overcome this limitation, recent research has turned its attention to RL~\citep{park2021schedulenet, zhang2020learning, kayhan2021reinforcement}, which can learn directly from simulations and observe performance without relying on manual labels.

In this report, we investigate the viability of RL for the JAP. We achieve this by formulating the problem within the RL framework and leveraging Deep Q-learning~\citep{mnih2015human} in conjunction with graph neural networks.
\section{The Job Allocation Problem}\label{sec:jsap}

In this section, we formally introduce the Job Allocation Problem (JAP), and its classical optimization formulation.

\textbf{Notation:} A graph $G(V, E)$ denotes a directed graph where $V$ represents the vertex set and $E$ represents the edge set. We note that $\{i, j\} \in E$ represents an undirected edge, while $(i, j) \in E$ represents a directed edge. We define $\mathcal{N}_{\text{in}}(v)$ and $\mathcal{N}_{\text{out}}(v)$ as the in- and out-neighborhoods of vertex $v$. The in- and out-degree of a vertex $v$ are then defined as $d_\text{in}(v) = \vert \mathcal{N}_\text{in}(v) \vert$ and $d_\text{out}(v) = \vert \mathcal{N}_\text{out}(v) \vert$.

\begin{definition}[Job Allocation Graph]
    An arbitrary graph $G(A, B)$ is called a job allocation graph if we can partition $G$ as $G(P \cup J, S \cup C)$ where $P \cap J = \emptyset$ and $S \cap C = \emptyset$, such that $\forall \{v, w\} \in S: [v \in P \land w \in J] \lor [v \in J \land w \in P]$ and $\forall (v, w) \in C: v, w \in J$.
    \label{def:instance}
\end{definition}

\begin{figure}[htb]
    \centering
    \includegraphics[width=0.35\linewidth]{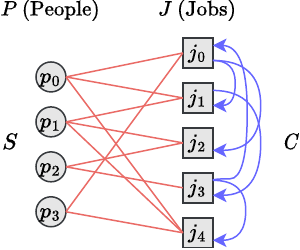}
    \caption{Example of the problem instance. We have individuals represented by $p_0, \cdots, p_3 \in P$, and jobs denoted by $j_0, \cdots, j_4 \in J$. The red selection edges from set $S$ connect people to jobs, signifying that a person is qualified to do a job. Additionally, there are directed blue conflict edges in set $C$. These connect jobs, indicating that if a person $p$ is assigned to a job vertex $j$, then that person cannot also be assigned to any $j^\prime \in \mathcal{N}_\text{out}(j)$.}
    \label{fig:example-graph}
\end{figure}

Whilst the remainder of this report will talk about people and jobs, any graph that satisfies \cref{def:instance} can be described under the framework introduced in this report. \cref{fig:example-graph} depicts an example of such a graph, where $P$ represent the people vertices and $J$ represents the job vertex set. The structure of the graph is bipartite between the people and job vertices, with additional edges from jobs to other jobs. We call the undirected edge set between jobs and people the "selection set" $S$. For directed edges from jobs to other jobs, we define the "conflict set" $C$. Formally, a directed edge $(j_i, j_k) \in C$ semantically means that if the job $j_i$ is selected by a person $p$, then the job $j_k$ cannot be done by the same person. An edge $(p, j) \in S$ means that person $p$ can do the job $j$.

\begin{definition}[Maximum Job Allocation]
    Given a graph $G(P \cup J, S \cup C)$ that adheres to \cref{def:instance}, a maximum job allocation is a solution to the following constrained maximization problem:
    \begin{equation*}
        \begin{aligned}
            \text{maximize } & \vert A \vert\;,\\
            \text{s.t. } & A \subseteq S \;,\\
            \text{and } & \forall \{p_i, j_i\}, \{p_i, j_k\} \in A: (j_i, j_k) \in C \lor (j_k, j_i) \in C \Rightarrow i = k\;.\\
        \end{aligned}
    \end{equation*}
    \label{def:maximum-alloc}
\end{definition}

We call a job allocation any subset $A \subseteq S$ that satisfies the constraints in \cref{def:maximum-alloc}. The goal of the JAP is to find the maximum number of job assignments such that there are no conflicts between the jobs assigned to the same person.
\section{Optimal Job Allocations through Reinforcement Learning}\label{sec:rl-formalism}

In \cref{sec:jsap}, we discuss how to formalize the JAP and how an optimum solution to the problem is defined. In this section, we discuss how we formulate the JAP as a sequential decision-making problem using the Reinforcement Learning framework.

\begin{definition}[Markov Decision Process (MDP)]
    A Markov Decision Process (MDP) is defined as a 5-tuple $(\mathcal{S}, \mathcal{A}, \mathcal{P}, \mathcal{R}, \gamma)$, where $\mathcal{S}$ is a set of states and $\mathcal{A}$ is a set of actions. $\mathcal{P}: \mathcal{S} \times \mathcal{A} \rightarrow \mathcal{S}$ is the transition function that captures the probability of transitioning to a state $s^\prime \in \mathcal{S}$ given a current state $s \in \mathcal{S}$ and an action $a \in \mathcal{A}$ taken in $s$. $\mathcal{R}: \mathcal{S} \times \mathcal{A} \rightarrow \mathcal{S}$ is a function that defines the reward given when transitioning to state $s^\prime \in \mathcal{S}$ from state $s \in \mathcal{S}$ and taking action $a \in \mathcal{A}$. Finally, $\gamma \in [0, 1]$ is the discount factor, which emphasizes the importance of future rewards.
    \label{def:mdp}
\end{definition}

When a problem can be formalized as an MDP, as seen in~\cref{def:mdp}, it can be solved using RL. We will begin by describing the state space $\mathcal{S}$, which will be the set of all job allocation graphs. This means that any job allocation graph is equivalent to a state in the MDP.

Given a job allocation graph, the actions should allow for a job assignment to be selected. The idea behind this approach is the fact that assignments can be selected sequentially while eliminating constraint violations with the transition function. Hence, the action space $\mathcal{A}$ for a job allocation graph $G(P \cup J, S \cup C)$ is equivalent to the selection set $S$, as these are all possible job assignments.

In order to ensure a valid job allocation graph is maintained after choosing an action, the transition function behaves as follows: given a graph $G(P \cup J, S \cup C)$ and action $\{p, j\}$, it deterministically maps to a new graph $G^\prime(P \cup J, S^\prime \cup C^\prime)$, where $S^\prime = S \setminus \{\{p, j\}\} \setminus \{\{p, j_i\} \vert (j, j_i) \in C\}$~\footnote{It is important to realize that, since job allocation graphs are equivalent to states, this represents a deterministic transition with $\mathbb{P}[G^\prime \vert G, \{p, j\}] = 1$.}. As can be seen, the transition function eliminates any assignments that person $p$ cannot do anymore after choosing to do job $j$. This ensures that any transition to a new graph maintains feasibility.

An episode with $T$ steps is completed when a terminal state is reached. A terminal state is a graph $G(P \cup J, S \cup C)$ where no further assignments are possible, that is, $S = \emptyset$. The set of all the job allocations in the episode $A = \{a_1, a_2, \cdots, a_T\}$ is a feasible solution to the problem. It is important to note that, starting from any job allocation graph, the optimum solution $A^*$ can be reached by picking the assignments in $A^*$ in any order as actions. One consideration is that this implies that there are a lot of symmetries, which can make it more difficult to solve the problem. This has been addressed in~\citet{kwon2020pomo}, but this work has not been adopted in this report.

The reward function gives a reward of $1$ at each step of the environment, which means that the maximum cumulative reward attainable can be bounded from above by $\vert S \vert$. The discount factor $\gamma$ is set to $1$ since the total cumulative reward from the beginning of the episode is equal to the cardinality of the set of solutions.

\begin{figure}[htb]
    \centering
    \includegraphics[width=0.7\linewidth]{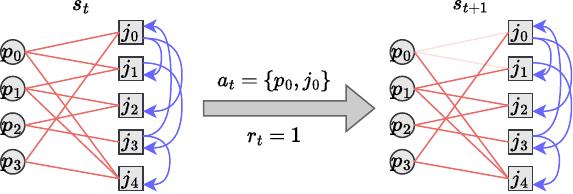}
    \caption{Example of a transition $(s_t, a_t, r_t, s_{t+1})$ in the MDP. When action $\{p_0, j_0\}$ is picked, its edge and the edge between $p_0$ and $j_1$, which conflicts with $j_0$, are removed from the graph in $s_{t+1}$ (depicted as transparent edges).}
    \label{fig:mdp}
\end{figure}

An example of a transition of the model can be seen in \cref{fig:mdp}. Since the state space $\mathcal{S}$ is too large for tabular reinforcement learning methods, we utilize techniques for reinforcement learning with function approximation. In~\cref{sec:methods}, we describe how we do this using graph neural networks.

\section{Graph Neural Network-Based Algorithm for MIS}
\label{sec:methods}

As discussed in \cref{sec:rl-formalism}, the state space of the MDP is too large to be approached with tabular reinforcement learning. By leveraging the graph structure of the states, we use graph neural networks to approximate $Q$-values. We first discuss the architecture of the model in~\cref{ssec:gnns}, and then describe the training pipeline in~\cref{ssec:training}.

\subsection{Function Approximation with Graph Neural Networks}
\label{ssec:gnns}

The goal is to design a model $Q_\theta(s, a) \approx Q^*(s, a)$, where $\theta \in \Theta$ are parameters and $Q^*(s, a)$ are the optimum $Q$-values. We develop a novel graph neural network module, the \textit{Context-Aware Embedding (CAE)} module, in order to effectively utilize the specific graph structure of the problem. We then use this module in the complete model architecture. 

\begin{figure}[htb]
    \centering
    \includegraphics[width=0.8\linewidth]{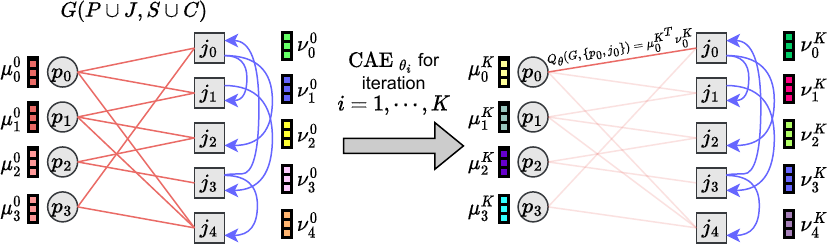}
    \caption{Overview of the model architecture. First, a job allocation graph $G(J \cup P, S \cup C)$ with initial node embeddings $\mu^0 \in \mathbb{R}^{\vert P \vert \times 2}, \nu^0 \in \mathbb{R}^{\vert J \vert \times 2}$ is put through $K$ Context-Aware Embedding modules with parameters $\theta_i \in \Theta$ for $i = 1, \cdots, K$. Afterward, $Q$-values can be predicted by doing an inner product of the corresponding vertex embeddings. For example, for the highlighted edge $\{p_0, j_0\}$, $Q_\theta(G, \{p_0, j_0\}) = \mu^{K^T}_0 \nu^K_0$.}
    \label{fig:gnn}
\end{figure}

Given a graph $G(J \cup P, S \cup C)$ and initial vertex embeddings $\mu^0 \in \mathbb{R}^{\vert P \vert \times 2}, \nu^0 \in \mathbb{R}^{\vert J \vert \times 2}$, which are just the out-degrees of the vertices for the two edge types, the high-level model architecture can be seen in~\cref{fig:gnn}. Initially, the graph and the embeddings are sequentially put through $K$ CAE modules, parameterized by $\theta_i \in \Theta$, where $\Theta$ for all $i = 1, \cdots, K$. We note that $\Theta$ is the parameter space. These modules compute node embeddings $\mu^i$ for person vertices and $\nu^i$ for job vertices for $i = 1, \cdots, K$. Finally, $Q$-values can easily be computed by taking a dot product between the final embeddings for two vertices of a given selection edge. For example, the estimated $Q$-value of choosing the person $p$ to do the job $j$ can be calculated as $Q_\theta(G, \{p, j\}) = \mu_p^{K^T} \nu_j^K$.

\begin{figure}[hbt]
    \centering
    \includegraphics[width=0.4\linewidth]{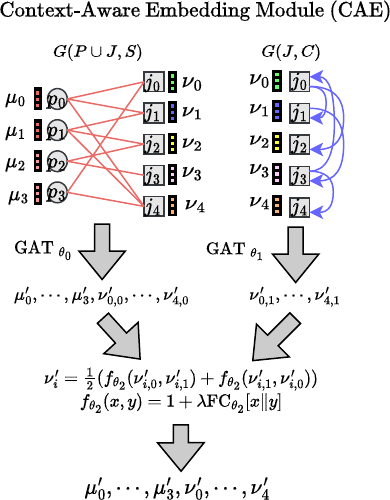}
    \caption{Overview of the Context-Aware Embedding Module. Given a graph $G(P \cup J, S \cup C)$ and vertex embeddings $\mu \in \mathbb{R}^{\vert P \vert \times d_{in}}, \nu \in \mathbb{R}^{\vert J \vert \times d_{in}}$, the module first splits it into two subgraphs $G(P \cup J, S)$ and $G(J, C)$. Then, these subgraphs are put through their own GAT layers, parameterized by $\theta_0, \theta_1 \in \Theta$. The final embeddings of the job vertices are then computed by combining them symmetrically using the learned linear function $f_{\theta_2}$ with parameters $\theta_2 \in \Theta$.}
    \label{fig:cae}
\end{figure}

\textbf{Context-Aware Embedding (CAE) Module:} 

The module is depicted in~\cref{fig:cae}. From a graph $G(J \cup P, S \cup C)$ and its current vertex embeddings $\mu \in \mathbb{R}^{\vert P \vert \times d_\text{in}}$ and $\nu \in \mathbb{R}^{\vert J \vert \times d_\text{in}}$ where $d_\text{in}$ is the dimensionality of a single vertex embedding. The CAE module operates by splitting $G$ into two subgraphs $G(J \cup P, S)$ and $G(J, C)$. Then, these graphs are put through their own attention layers~\citep{velickovic2018graph} with parameters $\theta_0, \theta_1 \in \Theta$ to update the node embeddings of the subgraphs. This splitting into subgraphs is inspired by~\citet{zhang2022learning}, which utilizes the same idea, since the edges have semantically different meanings. In the case of the JAP, selection edges are good to have, whilst conflict edges are bad to have. Since we have two sets of embeddings for the job vertictes after the attention layers, let's say $\nu^\prime_0 \in \mathbb{R}^{\vert P \vert \times d_\text{out}}$ and $\nu^\prime_1 \in \mathbb{R}^{\vert J \vert \times d_\text{out}}$ where $d_\text{out}$ is the output dimensionality of the vertex embeddings, we combine them into one as follows:

\begin{equation}
    \begin{aligned}
        \nu & = \frac{1}{2} \left(f_{\theta_2}(\nu_0, \nu_1) + f_{\theta_2}(\nu_1, \nu_0)\right)\;,\\
        f_{\theta_2}(x, y) & = 1 + \lambda \text{FC}_{\theta_2}(\left[x \Vert y\right])\;.
    \end{aligned}
    \label{eq:merger}
\end{equation}

Here, $\lambda \in \mathbb{R}$ is a learned parameter that modulates how far to deviate from a baseline value of $1$. Furthermore, $[x \Vert y] \in \mathbb{R}^{\vert x \vert + \vert y \vert}$ represents the vector concatenation of $x$ and $y$, and $\text{FC}_{\theta_2}$ is a fully connected layer with parameters $\theta_2 \in \Theta$. The form of this equation has been proposed by~\citet{BacMajMaoCotMcA20} and is meant to allow faster convergence. 

We note that, whilst it has been omitted from \cref{fig:gnn} for readability purposes, each CAE layer except for the last is followed by Layer Normalization~\citep{ba2016layer} and the GELU~\citep{hendrycks2016gaussian} activation function. 

\subsection{Optimizing the Model with Deep Reinforcement Learning}
\label{ssec:training}

We optimize the model using Double Deep Q-Learning~\citep{van2016deep}, which addresses stability and overestimation problems in Deep Q-Learning~\citep{mnih2015human}. In order to improve sample efficiency, we also utilize prioritized experience replay~\citep{schaul2015prioritized}, which samples from a replay buffer according to~\cref{eq:dist}. Here, $\delta_i$ represents the Q-learning error of sample $i$, and $\epsilon$ is a small value to avoid division by zero. Importance weights $w_i = \frac{(N \cdot P(i))^{-\beta}}{\max_j w_j}$ are used to correct the non-uniform sampling procedure. Hyperparameters $\alpha$ and $\beta$ regulate the sampling and correction strengths respectively.

\begin{equation}
    P(i) = \frac{p_i^\alpha}{\sum_j p_j^\alpha}\;,\;\; p_i = \vert \delta_i \vert + \epsilon\;.
    \label{eq:dist}
\end{equation}

The full algorithmic pipeline of the training process is described in~\cref{alg:train-pipeline}.

\begin{algorithm}[htb]
    \caption{Pipeline of the Training Approach}
    \label{alg:train-pipeline}
    \begin{algorithmic}[1]
    \State \textbf{Input:} A distribution $\mathcal{D}$ over job allocation graphs, soft update rate parameter $\tau$, prioritized experience replay parameters $\alpha$ and $\beta$, batch size $b$, learning rate $\eta$.
    \State Initialize $Q_\theta$ and make copy $Q_\text{target} \leftarrow Q_\theta$.
    \State Initialize a \textit{prioritized replay buffer} $\mathcal{B} \gets \varnothing$.
    \For{each episode $t=0,\ldots,T-1$}
    \State Sample a graph $G_{\mathrm{init}}(P \cup J, S \cup C) \sim \mathcal{D}$.
    \State Let $s \leftarrow G_{\mathrm{init}}(P \cup J, S \cup C)$.
    \While{$S \neq \emptyset$}
        \State Sample $a \leftarrow \{p \in P, j \in J\} \sim \text{softmax}_{\bar{a} \in S} ({Q_\theta(s, \bar{a})})$. \Comment{Sample a job assignment.}
        \State Receive $r \leftarrow 1$.
        \State Update $S^\prime = S \setminus \{\{p, j\}\} \setminus \{\{p, j_i\} \vert (j, j_i) \in C\}$. \Comment{Remove conflicting edges.}
        \State Update $s^\prime \leftarrow s(P \cup J, S^\prime \cup C)$.
        \State Store $(s, a, r, s^\prime, p)$ in $\mathcal{B}$.
        \State Set $\Delta = 0$.
        \For{$i = 1, \cdots, b$}
            \State Sample $k \sim \frac{p_k^\alpha}{\sum_j p_j^\alpha}$.
            \State Compute $w_k = \frac{(N \cdot P(k))^{-\beta}}{\max_j w_j}$.
            \State Compute $\delta_k = r_k + \gamma Q_{\text{target}}(s^\prime_k, \arg\max_{a^\prime} Q_\theta(s^\prime_k, a^\prime))  - Q_\theta(s_k, a_k)$.
            \State Update transition priority $p_k \leftarrow \vert \delta_i \vert + \epsilon$.
            \State Accumulate $\Delta \leftarrow \Delta + w_k \cdot \delta_k \cdot \nabla_\theta Q_\theta(s_k, a_k)$.
        \EndFor
        \State Update $\theta \leftarrow \theta + \eta\Delta$. \Comment{Mini-batch Gradient Descent.}
        \State Update $Q_{\text{target}} \leftarrow \tau Q_\theta + (1 - \tau) Q_{\text{target}}$.
        \State $s \leftarrow s^\prime$.
    \EndWhile
    \EndFor
    \end{algorithmic}
\end{algorithm}
\section{Experiments}
\label{sec:exp}

In this section, we evaluate the proposed approach of this report against baseline algorithms. First, we describe the training setup and datasets. Additional experiments on insights into the learned representations of the model can be found in \cref{app:add-exp}.

\subsection{Training Setup}

\textbf{Our model:} We use the model with $16$ neurons in all hidden layers, and $8$ in the output layer. Each hidden layer is equipped with Layer Normalization, followed by the GELU activation function. We stack $3$ CAE modules, thus setting $K = 3$ in~\cref{fig:gnn}. We refer to this model as the \textit{GNN}.

The specific hyperparameters during training can be found in \cref{app:exp-details}. The model is optimized for $200$ episodes using the Adam optimizer with weight decay~\citep{loshchilov2017decoupled}.

\textbf{Baselines:} We utilize three baseline algorithms to compare our performance. The first, which we refer to add \textit{Greedy}, first selects the job with the minimum degree. Then, for that job, it chooses the person with the minimum degree connected to it. We also have a \textit{Random} algorithm, which randomly selects assignments and an \textit{Untrained GNN} with a random model initialization.

\textbf{Datasets:} We evaluate our model on a real-world hospital dataset (\textit{Planny}) and two synthetically generated datasets. The latter two are random network datasets generated using the \textit{Erdős–Rényi} and \textit{Barabási–Albert} models. Some key statistics of these datasets can be found in \cref{app:exp-details}.

\subsection{Results}

As per the findings presented in \cref{tab:benchmark}, it becomes evident that the Greedy algorithm serves as a robust baseline across all datasets. However, the Graph Neural Network (GNN) demonstrates superior performance across each dataset, particularly for Erdős–Rényi instances. Nevertheless, for the Barabási–Albert dataset, it remains uncertain if the model has effectively acquired any meaningful knowledge, given that the minimum performance is already nearly optimal. 

\begin{table}[h]
    \centering
    \caption{Approximation ratios (higher is better; the best performance in bold) on different synthetic datasets and a real-world dataset (Planny). We report the average approximation ratios along with the standard deviation.}
    \begin{tabular}{|c|c|c|c|}
        \hline
        Method ($\downarrow$) Dataset ($\rightarrow$) & Planny & Erdős–Rényi & Barabási–Albert\\ \hline
        GNN & $\mathbf{0.989 \pm 0.013}$ & $\mathbf{0.981 \pm 0.010}$ & $0.999 \pm 0.002$\\\hline
        Greedy & $0.987 \pm 0.026$ & $0.962 \pm 0.014$ & $\mathbf{1.000}$\\ \hline
        Random & $0.969 \pm 0.041$ & $0.924 \pm 0.021$ & $0.996 \pm 0.004$ \\\hline
        Untrained GNN & $0.916 \pm 0.088$ & $0.915 \pm 0.027$ & $0.999 \pm 0.005$ \\\hline
    \end{tabular}
    \label{tab:benchmark}
\end{table}

To assess the generalizability of each model, out-of-distribution testing is conducted on the remaining datasets, and the corresponding results are provided in \cref{tab:ood}. The model trained on the Barabási–Albert dataset exhibits performance on par with a random model, suggesting that this dataset might not have provided the model with ample opportunities to learn a valuable representation, given that most assignments are already close to optimality. Interestingly, the other models achieve optimal scores on the Barabási–Albert dataset. Both the Planny GNN and the Erdős–Rényi GNN surpass the Greedy algorithm on the other datasets, despite not being trained on them.

\begin{table}[h]
    \centering
    \caption{Approximation ratios (higher is better; the best performance in bold) of out-of-distribution models. We report the average approximation ratios along with the standard deviation.}
    \begin{tabular}{|c|c|c|c|}
        \hline
        Method ($\downarrow$) Dataset ($\rightarrow$) & Planny & Erdős–Rényi & Barabási–Albert\\ \hline
        Planny GNN & $\mathbf{0.989 \pm 0.013}$ & $0.970 \pm 0.012$ & $\mathbf{1.000}$\\
        Erdős–Rényi GNN & $0.984 \pm 0.023$ & $\mathbf{0.981 \pm 0.010}$ & $\mathbf{1.000}$\\
        Barabási–Albert GNN & $0.968 \pm 0.040$ & $0.928 \pm 0.024$ & $0.999 \pm 0.002$\\
        \hline
    \end{tabular}
    \label{tab:ood}
\end{table}

\begin{figure}[htb]
    \centering
    \subfloat[Changing the probability of an edge with a fixed $300$ jobs. It was trained on graphs with an edge probability of $0.10$.]{%
        \includegraphics[width=0.45\textwidth]{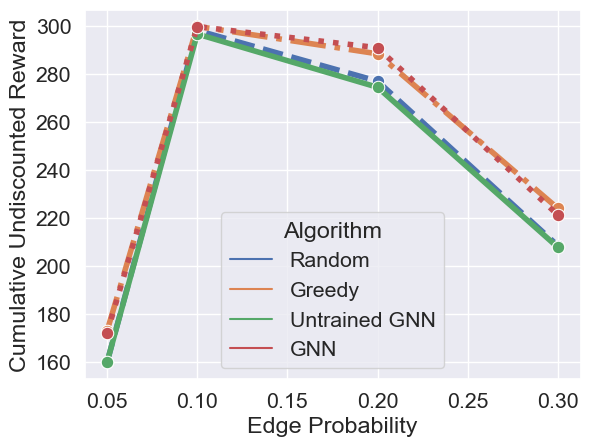}%
        \label{fig:subfig1}%
    }\hfil 
    \subfloat[Changing the number of jobs with a fixed edge probability of $0.10$. It was trained on graphs with $300$ jobs.]{%
        \includegraphics[width=0.45\textwidth]{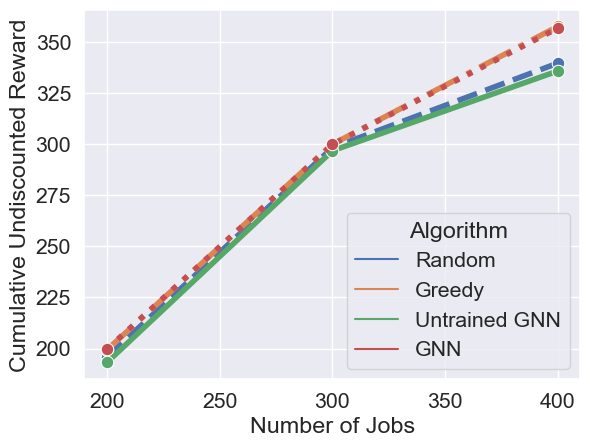}%
        \label{fig:subfig2}%
    }
    \caption{Out-of-distribution performance of the algorithms when tweaking the individual parameters of the Erdős–Rényi model.}
    \label{fig:ood-er}
\end{figure}

Finally, we conduct out-of-distribution experiments using datasets generated by Erdős–Rényi models, varying the numbers of jobs and densities of graphs. The experimental results are illustrated in \cref{fig:ood-er}. Our findings demonstrate that the performance of the GNN model is comparable to, and occasionally surpasses, that of the Greedy model. These outcomes suggest favorable scalability of the GNN model to larger and more diverse instances.

\section{Conclusion}
\label{sec:conc}
In this report, we proposed a novel approach that combines Reinforcement Learning with Graph Neural Networks to tackle the Job Allocation Problem. We formulated the JAP as a Markov Decision Process and developed a model architecture that utilizes Graph Neural Networks to approximate Q-values. Our approach eliminates the need for manual annotation, which is often a major bottleneck in supervised learning approaches.
Through experimental evaluations on real-world and synthetic datasets, we demonstrated that our proposed approach outperforms baseline algorithms, including a greedy algorithm and random models. The Graph Neural Network consistently achieves higher approximation ratios, showcasing its effectiveness in solving the JAP. Furthermore, the GNN model exhibited generalization capabilities, achieving competitive performance on out-of-distribution datasets. Overall, our findings highlight the potential of leveraging Reinforcement Learning and Graph Neural Networks for optimizing job allocation in complex scheduling problems. Future work can explore further improvements to the model architecture, as well as investigate its application in other real-world domains and more complex and constrained problem settings.


\bibliography{bib}
\bibliographystyle{plainnat}

\newpage
\appendix

\section*{Contents of the appendix}

We describe the contents of the supplementary materials below:

\begin{itemize}
    \item In \cref{app:exp-details}, we discuss the specific hyperparameters that were used for the training of the models. We also describe some of the key statistics of the datasets.
    \item In \cref{app:add-exp}, we discuss additional experiments performed with the framework regarding looking into the learned representation of the model.
\end{itemize}

\section{Experimental Details}
\label{app:exp-details}

For the experiments in \cref{sec:exp}, the hyperparameters described in \cref{tab:hyperparameters} are used. Furthermore, the most important statistics of the datasets are described in \cref{tab:data-stats}.

\begin{table}[htb]
    \centering
    \caption{Hyperparameters employed in the training process.}
    \begin{tabular}{|c|c|}
         \hline
         Name & Value\\
         \hline
         Learning rate & $0.001$ \\
         Optimizer & AdamW \\
         Hidden layer size & $16$\\
         Output layer size & $8$ \\
         Number of GEM layers (K) & $3$\\
         Batch size & $2048$\\
         Number of episodes & $200$\\
         $\epsilon$ (epsilon-greedy) & $0.10$\\
         $\gamma$ (discount factor) & $1.0$\\
         $\theta$ (soft update rate) & $0.025$\\
         Replay memory size & $10^6$\\
         $\alpha$ (PER) & $0.6$\\
         $\beta$ (PER) & $0.4$\\     
         \hline
    \end{tabular}
    \label{tab:hyperparameters}
\end{table}

\begin{table}[htb]
    \centering
    \caption{Key statistics about the datasets.}
    \resizebox{\textwidth}{!}{%
    \begin{tabular}{|c|c|c|c|c|c|c|}
         \hline
         Dataset Name & \#Graphs & Mean \#jobs & Mean \#people & Mean \#job conflicts & Mean \#assignments & Mean density\\
         \hline
         Planny & 20 & 507.2 & 25.2 & 14344.5 & 5777.9 & 0.079\\
         \hline
         Erdős–Rényi & 20 & 300.0 & 15.0 & 8998.5 & 2998.75 & 0.152\\
         \hline
         Barabási–Albert & 20 & 300.0 & 15.0 & 1782.0 & 3137.65 & 0.081\\
         \hline
    \end{tabular}%
    }
    \label{tab:data-stats}
\end{table}

\section{Additional Experiments with Learned Model Representations}
\label{app:add-exp}

\begin{wrapfigure}{r}{5cm}
    \centering
    \includegraphics[width=\linewidth]{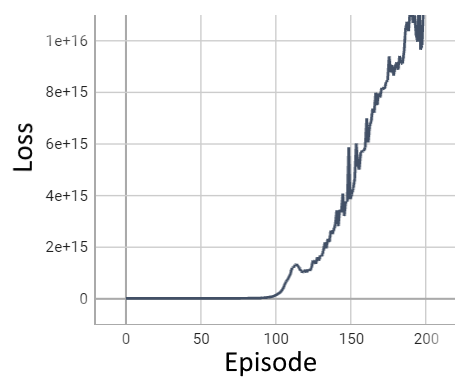}
    \caption{Exploding loss over the training of a GNN agent.}
    \label{fig:loss}
\end{wrapfigure}

In order to gain insights into the results of the model after training, we look at two figures. \cref{fig:loss} shows that the loss explodes during training. This is not surprising, as a reward of $1$ at every step can lead to instability over training as many less terminal states are encountered. Since terminal states are the only states where the model can observe that the reward doesn't just always happens, it is expected that the Q-values can explode. In order to address this instability, Double Deep Q-Learning and Prioritized Experience Replay (PER) were utilized as described in \cref{sec:methods}. Resulting from the performance of the model in \cref{sec:exp}, it can be concluded that the model was still able to learn a good order of states with the learned Q-values, even though they explode. 

\begin{figure}[hbt]
    \centering
    \includegraphics[width=0.5\linewidth]{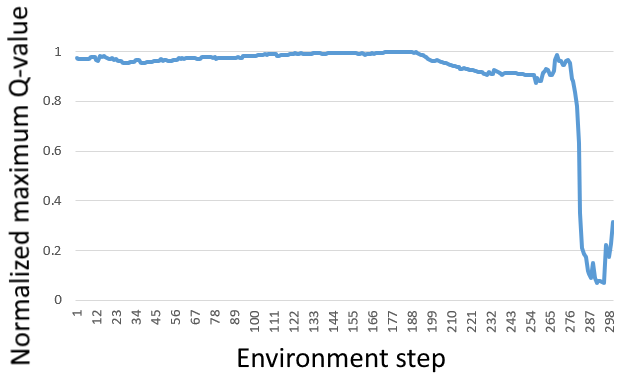}
    \caption{Maximum normalized estimated Q-values from a trained GNN agent over sequential timesteps of an episode.}
    \label{fig:q-vals}
\end{figure}

In order to confirm this idea, \cref{fig:q-vals} plots the estimated Q-values of a trained GNN divided by the maximum estimated Q-value over one episode. Since the state graphs in the episode should get smaller at each step, it is expected that the Q-values should decrease. From \cref{fig:q-vals}, it becomes apparent that the model does learn some decrease in these Q-values, although it does not seem decreasing monotonically.

\end{document}